\documentclass[journal]{IEEEtran}
\usepackage{amsmath,amsfonts}
\usepackage{algorithmic}
\usepackage{algorithm}
\usepackage{array}
\usepackage[caption=false,font=normalsize,labelfont=sf,textfont=sf]{subfig}
\usepackage{textcomp}
\usepackage{stfloats}
\usepackage{url}
\usepackage{orcidlink}
\usepackage{orcidlink}
\usepackage{tikz}
\usepackage{verbatim}
\usepackage{graphicx}
\usepackage{multirow}
\usepackage{booktabs}
\usepackage{biblatex}
\usepackage[T1]{fontenc}
\def\BibTeX{{\rm B\kern-.05em{\sc i\kern-.025em b}\kern-.08em T\kern-.1667em\lower.7ex\hbox{E}\kern-.125emX}}
\usepackage{balance}
\hypersetup{
colorlinks=true,
linkcolor=black,
citecolor=black
}

\begin{document}
\title{PMMTalk$:$ Speech-Driven 3D Facial Animation from Complementary Pseudo Multi-modal Features}
\author{Tianshun Han~\orcidlink{0009-0004-3393-1597}, Shengnan Gui, Yiqing Huang, Baihui Li,  

Lijian Liu, Benjia Zhou~\orcidlink{0000-0003-4883-5552}, Ning Jiang, Quan Lu, Ruicong Zhi,

Yanyan Liang~\orcidlink{0000-0002-5780-8540},~\IEEEmembership{Member,~IEEE}, Du Zhang,~\IEEEmembership{Senior Member,~IEEE}, Jun Wan~\orcidlink{0000-0002-4735-2885},~\IEEEmembership{Senior Member,~IEEE}
\thanks{
Tianshun Han, Yiqing Huang, Benjia Zhou, Yanyan Liang and Du Zhang are with the School of Computer Science and Engineering, the Faculty of Innovation Engineering, Macau University of Science and Technology, Macau 999078, China. (e-mail: \{3230002542, 3230006115, 21098536ia30001\}@student.must.edu.mo, \{yyliang, duzhang\}@must.edu.mo).

Baihui Li, Lijian Liu and Jun Wan are with the State Key Laboratory of Multimodal Artificial Intelligence Systems (MAIS), Institute of Automation, Chinese Academy of Sciences (CASIA), Beijing 100190, China, also with the School of Artificial Intelligence, University of Chinese Academy of Sciences (UCAS), Beijing 100049, China. Jun Wan is also with the School of Computer Science and Engineering, the Faculty of Innovation Engineering, Macau University of Science and Technology, Macau 999078, China. (e-mail: \{libaihui2022, liulijian2022, jun.wan\}@ia.ac.cn)

Shengnan Gui and Ruicong Zhi are with the School of Computer and Communication Engineering, University of Science and Technology Beijing, Beijing 100083, China. Ruicong Zhi is also with the Beijing Key Laboratory of Knowledge Engineering for Materials Science, Beijing 100083, China. (e-mail: m202110625@xs.ustb.edu.cn; zhirc@ustb.edu.cn)

Ning Jiang and Quan Lu are with the Mashang Consumer Finance Co., Ltd, Chongqing 400000, China. (e-mail: ning.jiang02@msxf.com; quan.lu@gmail.com)
}}

\markboth{}{PMMTalk: Speech-Driven 3D Facial Animation from Complementary Pseudo Multi-modal Features}
\maketitle 
\begin{abstract}
Speech-driven 3D facial animation has improved a lot recently while most related works only utilize acoustic modality and neglect the influence of visual and textual cues, leading to unsatisfactory results in terms of precision and coherence. We argue that visual and textual cues are not trivial information. Therefore, we present a novel framework, namely PMMTalk, using complementary \textbf{P}seudo \textbf{M}ulti-\textbf{M}odal features for improving the accuracy of facial animation. The framework entails three modules: PMMTalk encoder, cross-modal alignment module, and PMMTalk decoder. Specifically, the PMMTalk encoder employs the off-the-shelf talking head generation architecture and speech recognition technology to extract visual and textual information from speech, respectively. Subsequently, the cross-modal alignment module aligns the audio-image-text features at temporal and semantic levels. Then PMMTalk decoder is employed to predict lip-syncing facial blendshape coefficients. Contrary to prior methods, PMMTalk only requires an additional random reference face image but yields more accurate results. Additionally, it is artist-friendly as it seamlessly integrates into standard animation production workflows by introducing facial blendshape coefficients. Finally, given the scarcity of 3D talking face datasets, we introduce a large-scale \textbf{3D} \textbf{C}hinese \textbf{A}udio-\textbf{V}isual \textbf{F}acial \textbf{A}nimation (3D-CAVFA) dataset. Extensive experiments and user studies show that our approach outperforms the state of the art.\footnote{We will release the code and dataset when this paper is accepted.} We recommend watching the supplementary video.
\end{abstract}

\begin{figure}[htbp]
\centering
\includegraphics[width=8.7cm]{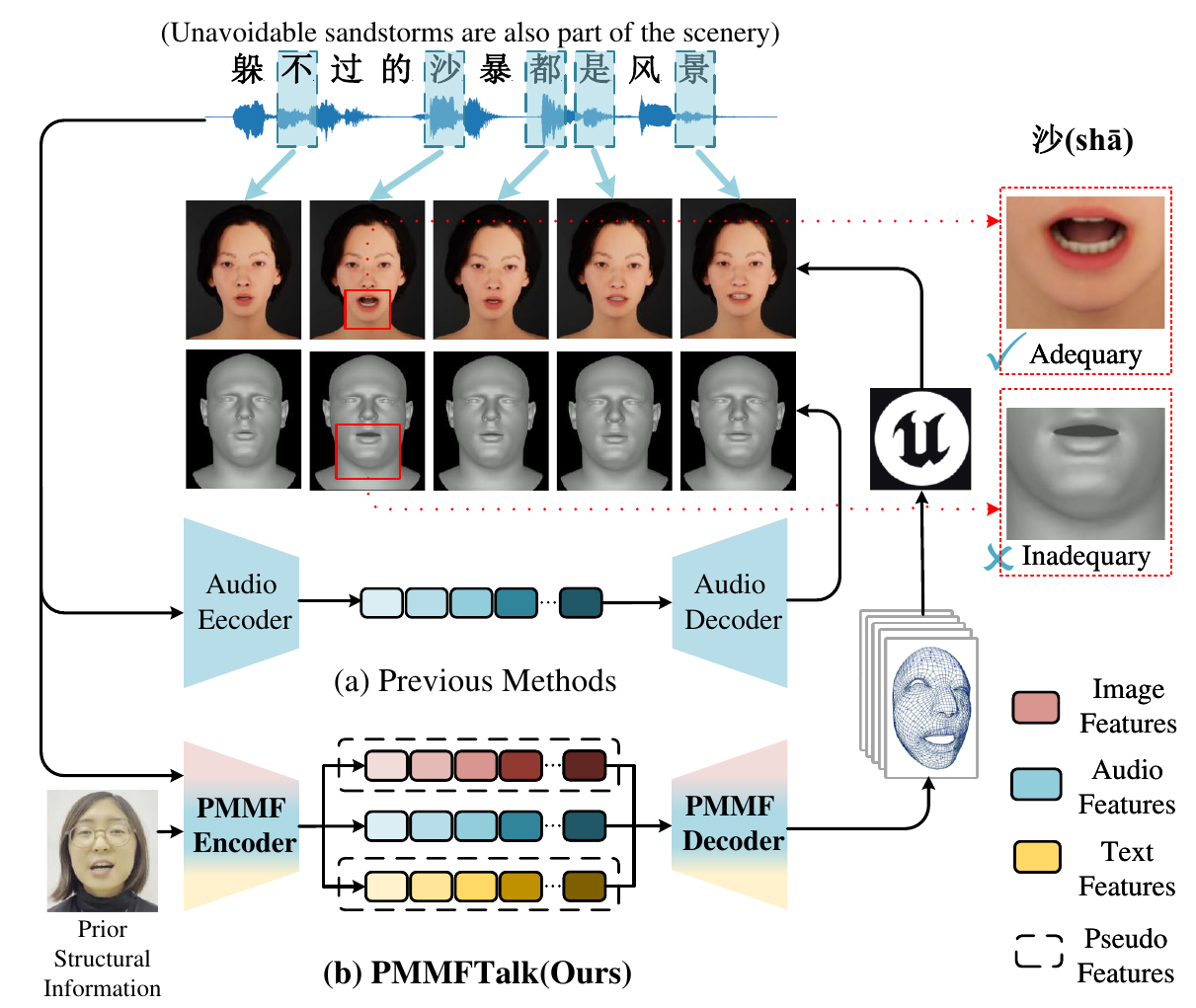}
\caption{\textbf{Comparison with existing methods.} (a) Prior methods relying solely on acoustic features result in inadequate and insufficient lip movements such as in the pronunciation of Chinese word $/sha/$, demanding a full mouth opening and downward jaw movement. (b) PMMTalk utilizes complementary pseudo multi-modal features generated from speech, enhancing the accuracy of facial animation. In addition, unlike previous methods that could only generate vertex-based animation, PMMTalk predicts synchronized facial blendshape coefficients, which is more artist-friendly.}
\label{concept—diagram}
\end{figure}

\section{Introduction}
Speech-driven 3D facial animation for digital characters has gained popularity in academia and industry, offering potential benefits for virtual reality~\cite{fan2022object,fan2022deep}, film production~\cite{ye2022perceiving,chen2009analysis}, and computer gaming~\cite{ping2013computer,fan2022reconstruction}. Presently, commercial 3D facial blendshape creation by animators is laborious and time-consuming. Hence, animators are seeking an artist-friendly approach that can both effectively reduce expenses and seamlessly integrate into existing workflows~\cite{bao2023learning,zhou2018visemenet}.

Numerous  methods~\cite{karras2017audio,voca,fan2022faceformer,xing2023codetalker,haque2023facexhubert,peng2023selftalk,emotalk} for speech-driven 3D facial animation have been proposed. VOCA~\cite{voca} uses time convolutions to regress the face movement from audio, and FaceFormer~\cite{fan2022faceformer} presents a transformer-based autoregressive architecture for generating continuous facial movements. However, these methods face two critical issues. In one aspect, they only utilize audio signals (as depicted in Fig.~\ref{concept—diagram}), disregarding potent visual and textual cues. Speech signals are ambiguous as some phoneme pairs can be easily confusable on the basis of acoustics alone. Actually, the human speech perception inherently involves acoustic, visual, and textual features~~\cite{dupont2000audio,fan2022joint,mcgurk1976hearing}. For instance, visual cues like jaw and lower face muscle movements have a correlation with pronunciation~\cite{yehia1998quantitative}, serving to assist in disambiguation~\cite{potamianos2004audio}, while textual cues can enhance the comprehension and understanding of models by providing diverse contextual information and semantic properties~\cite{fan2022joint}. In another aspect, as shown in Table~\ref{datasets}, existing 3D talking face datasets~\cite{BIWI,voca,meshtalk,emotalk,wu2023mmface4d} have certain issues: (1) \textbf{Scarcity.} Speech signals and facial motion exhibit strong correlations, yet they exist in distinct spaces, necessitating a substantial amount of training data~\cite{voca,wang20213d}. However, only a limited number of publicly available datasets exist, and they suffer from limitations in terms of scale. (2) \textbf{Diversity.}  Most existing datasets narrow their focus to a single corpus type, such as sentences, failing to capture the rich diversity of language usage patterns that are encountered in real-world scenarios. (3) \textbf{Generic.} The majority of existing datasets employ vertex offsets of mesh as labels, which are not artist-friendly and pose challenges in integrating models based on these datasets into typical animation production workflows~\cite{bao2023learning}. 

To address the first challenge, we present a novel framework, namely PMMTalk (as illustrated in Fig.~\ref{method}), utilizing complementary \textbf{P}seudo \textbf{M}ulti-\textbf{M}odal features for enhancing the accuracy of facial animation. The framework consists of three modules: PMMTalk encoder, cross-modal alignment module, and PMMTalk decoder. Specifically, the PMMTalk encoder uses the off-the-shelf talking head generation method and speech recognition technology to extract visual and textual information from speech, respectively. The cross-modal alignment module then aligns these audio-image-text features to work together on both temporal and semantic levels. The PMMTalk decoder driven by the audio features, image features, text features, and personal style embeddings predicts lip-syncing facial blendshape coefficients while controlling personal styles. To cope with the second aspect, we public a large-scale \textbf{3D} \textbf{C}hinese \textbf{A}udio-\textbf{V}isual \textbf{F}acial \textbf{A}nimation (\textbf{3D-CAVFA}) dataset, comprising 20 subjects with synchronized facial blendshape coefficients, and a total duration of almost 15.3 hours. It covers frequently used single words and phrases in modern Chinese, and encompasses various commonly used sentence patterns in daily life, making it highly relevant to everyday situations. Finally, we propose two evaluation protocols to assess model performance from different perspectives, including cross-subject and cross-gender ability, bridging the gap between academic research and real-world applications. The contributions of our work are as follows:

\begin{itemize}
\item To the best of our knowledge, we are the pioneers in leveraging pseudo multi-modal information into speech-driven 3D facial animation, outperforming existing state-of-the-art methods.
\\
\item We propose a novel and effective architecture, namely \textbf{PMMTalk}, which extracts pseudo visual and textual information from speech and aligns the audio-image-text features at temporal and semantic levels, improving the accuracy of facial animation.
\\
\item We present a large-scale 3D talking face dataset, namely the \textbf{3D-CAVFA} dataset, with synchronized facial blendshape coefficients, diverse corpus, and various subjects. 
\end{itemize}  

\section{Related Work}
\label{sec:related}
\begin{table*}[!htbp]
\begin{center}
\caption{\textbf{Comparison with publicly available 3D talking face datasets.} The 3D-CAVFA dataset stands out regarding subject numbers(\#Subj), text corpus content(\#Crop), and duration(\#Dura). We also list public year(\#Year), frame per second(\#FPS), spoken language(\#Lang), and Labels(\#Lab).} 
\renewcommand\arraystretch{1.6} 
  \begin{tabular}{c|c|c|c|c|c|c|c}
   \cline{1-8}
    \makebox[0.12\textwidth][c]{Dataset} & \makebox[0.04\textwidth][c]{\#Year} & 
    \makebox[0.04\textwidth][c]{\#Subj}  & \makebox[0.06\textwidth][c]{\#Dura} & 
    \makebox[0.10\textwidth][c]{\#Crop}  & \makebox[0.04\textwidth][c]{\#FPS}  &
    \makebox[0.06\textwidth][c]{\#Lang}  & \makebox[0.10\textwidth][c]{\#Lab} \\
    \cline{1-8}
    BIWI~\cite{BIWI}         & 2010 & 14  & 1.44h & Sentences                    & 25 & English & Vertex Positions\\
    VOCASET~\cite{voca}      & 2019 & 12  & 0.5h  & Sentences,Pangrams,Questions & 60 & English & Vertex Positions\\
    MeshTalk~\cite{meshtalk} & 2021 & 250 & 13h   & Sentences                    & 30 & English & Vertex Positions\\
    \cline{1-8}
    EmoTalk~\cite{emotalk}   & 2023 & 12  & 6.5h  & Sentences                    & 30 & English & Blendshape Coefficients\\
    3D-CAVFA                & - & 20  & 15.3h & Words,Phrases,Sentences      & 60 & Chinese & Blendshape Coefficients\\
    \cline{1-8}
  \end{tabular}
\label{datasets}
\end{center}
\end{table*}
\subsection{Speech-Driven 3D Facial Animation}
Formerly, there are numerous studies focusing on 2D talking head generation~\cite{,wang2022anyonenet,chen2019hierarchical,,prajwal2020lip,ji2021audio,hong2022depth,ye2022audio,song2022audio,thies2020neural,eskimez2021speech}, utilizing either image-driven or speech-driven approaches to create realistic videos of individuals speaking. Nevertheless, these approaches are not suitable for 3D character models, which are extensively utilized in 3D games and virtual reality interactions. Therefore, we focus on speech-driven 3D facial animation~\cite{karras2017audio,voca,richard2021meshtalk,fan2022faceformer,emotalk} in this work.

Karras et al.~\cite{karras2017audio} leverages an end-to-end convolutional network for mapping speech features to 3D vertex positions and designs an emotion states code to explain various facial expressions. VOCA~\cite{voca} utilizes advanced audio feature extraction models and has the capability to generate facial animation with diverse speaking styles. MeshTalk~\cite{richard2021meshtalk} focuses on the upper part of the face, learning a categorical latent space that successfully disentangles audio-correlated and audio-uncorrelated face motions. FaceFormer~\cite{fan2022faceformer} employs a self-supervised pre-trained speech model to extract audio features, and then put them into a transformer-based model and autoregressively generate continuous facial animations. Most related to our work is EmoTalk~\cite{emotalk}, whereby proposing an emotion disentangling encoder to disentangle the emotion as well as content in the speech and output emotional blendshape coefficients. 

However, they ignore the strong influence between acoustic, visual, and textual cues, limiting the accuracy of lip movements and causing ambiguity for identical pronunciations. Additionally, most of these methods are primarily focused on directly generating vertex-based animation from audio, which poses challenges in terms of user-friendliness for artists and integration into existing animation production workflows. Therefore, in this paper, we present an effective architecture utilizing complementary pseudo multi-modal features for improving the accuracy of facial animation.

\subsection{Talking Face Dataset}
Several 3D~\cite{cao2013facewarehouse,paysan20093d,savran2008bosphorus,yang2020facescape} and 4D face datasets~\cite{alashkar20143d,cosker2011facs,zhang2013high,zhang2014bp4d,zhang2016multimodal,fanelli20103,4DFAB} have been published for both static and dynamic facial expression recognition, while only a limited number of datasets~\cite{BIWI,voca,meshtalk,emotalk} available that capturing synchronized audio along with dynamic 3D facial movements. As shown in Table~\ref{datasets}, we have curated a collection of currently available open-source datasets.

BIWI~\cite{BIWI} consists of 40 spoken English sentences for each of the 14 subjects, while VOCASET~\cite{voca} captures 29 minutes of 4D scans from 12 speakers. However, the limited size of both the BIWI and VOCASET datasets restricts the generalization capacity of 3D face animation models. MeshTalk~\cite{meshtalk} collects 250 subjects each reading 50 sentences, while it only publishes a few parts of the dataset. Recently, EmoTalk~\cite{emotalk} publishes a large-scale 3D emotional talking face dataset by predicting the facial blendshape coefficients from the corresponding input 2D video streams. 

However, they face issues of restricted corpus diversity and dataset scale, which consequently limits the ability of models to comprehend complex relations between speech and lip movements. Further, the majority utilize vertex offsets of mesh as labels, imposing a limitation on the generalization ability of models trained on these datasets~\cite{emotalk}. To overcome these limitations, we collect a large-scale 3D talking face dataset, namely the 3D-CAVFA dataset, characterized by synchronized facial blendshape coefficients, diversified corpus, and assorted subjects.


\section{Datasets construction}
As commented, all existing datasets face some limitations, such as restricted corpus diversity, limited dataset scale, and a reduced number of subjects, impeding the development of speech-driven 3D facial animation technology. Conventionally, these datasets adopt vertex offsets of mesh as labels, which inadvertently imposes limitations on the generalization ability of models trained on these resources. To mitigate these issues and advance the field, we assemble a large-scale 3D Chinese audio-visual facial animation dataset, namely the 3D-CAVFA dataset. This dataset, comprising approximately 3,150,000 frames of facial blendshape coefficients over a span of 15 hours, presents a significant improvement over previous resources. As such, our initiative addresses a notable paucity in the availability of large-scale 3D facial animation datasets. We are preparing to make the 3D-CAVFA dataset accessible to the academic community in the near future.

\begin{figure}[htbp]
\centering
\includegraphics[width=8.7cm]{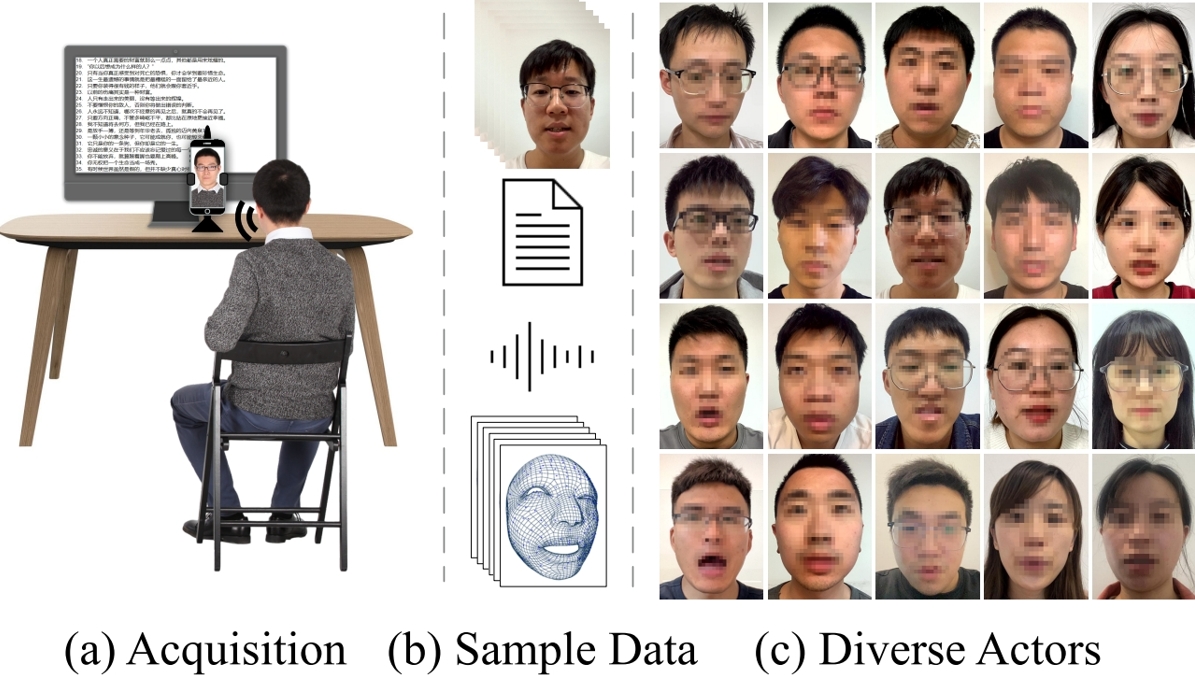}
\caption{\textbf{Overview of proposed 3D-CAVFA dataset.} \textbf{(1) Acquisition:} We detail the setup and recording conditions used in data collection. \textbf{(2) Sample data:} A specific segment is showcased, including RGB videos, corresponding text information, synchronized speech audio, and facial blendshape coefficients. \textbf{(3) Diverse actors:} The dataset encompasses a diverse range of actors, further enhancing its representation and applicability. All RGB facial images are mosaicked for privacy.}
\label{dataset}
\end{figure}

\textbf{Corpus.} Our text corpus, aimed at maximizing phonetic diversity, contains Chinese main components and is divided into two segments. The first segment is derived from FACEGOOD\footnote{https://github.com/FACEGOOD/FACEGOOD-Audio2Face}, comprising single words and phrases that are frequently utilized in modern Chinese. The second segment is sourced from celebrated movie dialogues, both local and international, encapsulating a wide variety of sentence patterns commonly used in daily conversation. Therefore, our curated text corpus is designed to encompass an equitable distribution of each phoneme to the greatest extent feasible.

\textbf{Scene.} As illustrated in Fig.~\ref{dataset}, facial video recordings are conducted in a conference room with a plain white background. We employ the mobile software Live Link Face\footnote{https://apps.apple.com/us/app/live-link-face/id1495370836}, a free IOS application, to capture the facial movements of volunteers while speaking. For the assurance of capturing distinct facial details, we meticulously position the iPhone at face level, maintaining a distance of 0.5 meters. In comparison to traditional enterprise-level capture systems, this method boasts both superior scene adaptability and decreased hardware requirements, resulting in a cost-effective solution that still guarantees the capture of high-quality data.

\textbf{Actors.} We invite 20 student volunteers, composed of 6 females and 14 males, all of whom are native Mandarin speakers from China. To mimic real-life scenarios, no restrictions are placed on the speaking speed or duration. Participants are encouraged to read the text corpus naturally, and minor errors, such as the usage of synonyms that do not alter the intended meaning, are overlooked.

\textbf{Statistics.} The software records facial blendshape coefficients in 61 dimensions synchronized with timestamps. It captures information at a rate of 60fps and a resolution of 1080P to guarantee the seamless reproduction of facial movements. Simultaneously, the audio is sampled at a high rate of 48kHz to uphold superior sound quality. 

\textbf{Evaluation protocol.} To provide a comprehensive evaluation of model capabilities, we have devised two distinctive testing protocols: the cross-subject protocol and the cross-gender protocol. With the cross-subject protocol, the dataset is randomly partitioned: 60\% is designated for training, 20\% for validation, and a further 20\% reserved for testing, all based on the identity of the subjects. Within the cross-gender protocol, we employ all male samples for training purposes, while dividing the female samples evenly across the validation and test sets according to the the identity of the subjects.

\section{Method}
\begin{figure*}[htbp]
\centering
\includegraphics[width=18cm]{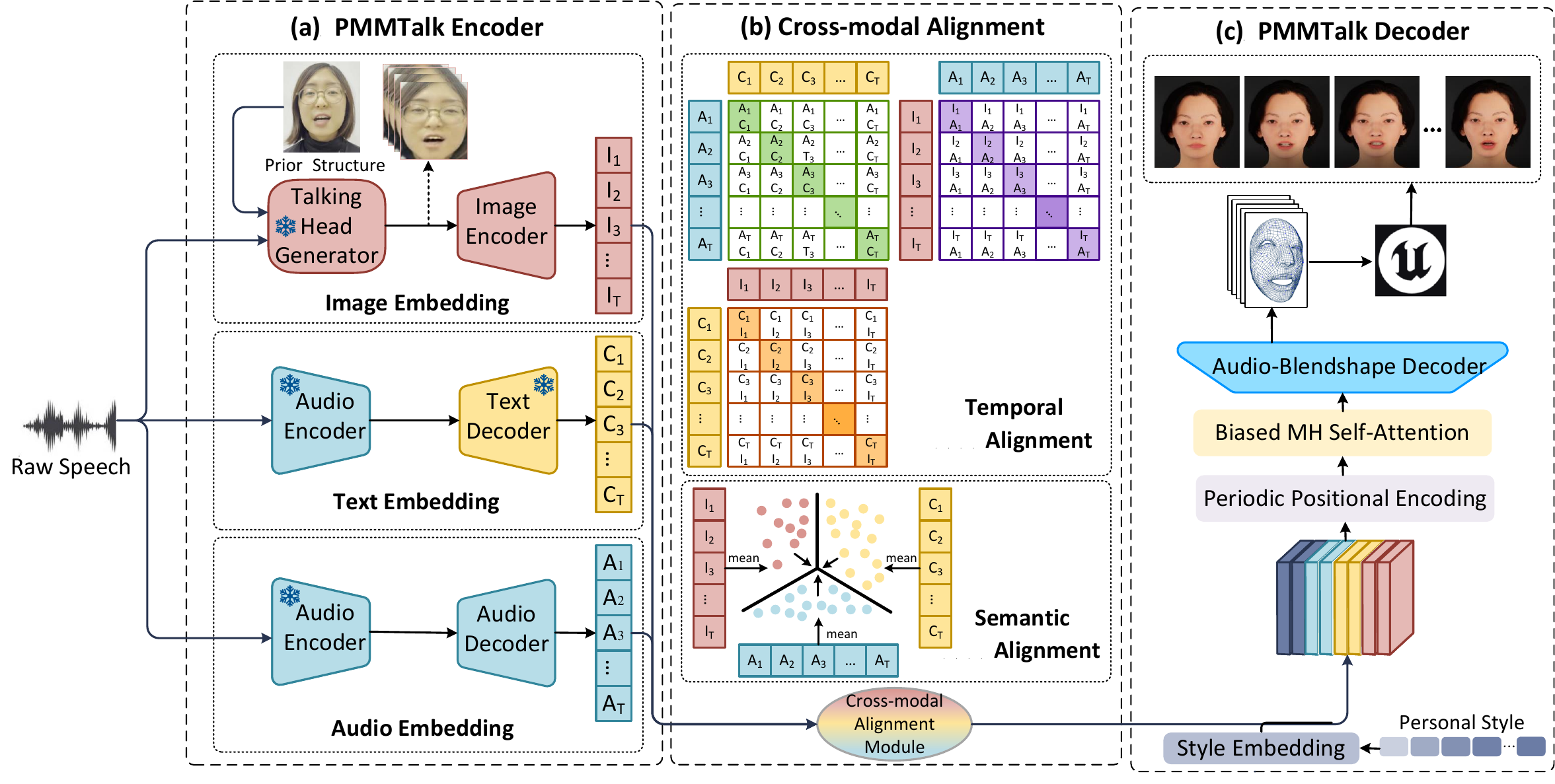}
\caption{\textbf{Overview of PMMTalk.} The framework consists of three modules: PMMTalk encoder, cross-modal alignment module, and PMMTalk decoder. (a) Given a raw speech $\mathcal{X}$, a random reference face image $R$ as inputs, the PMMTalk encoder can extract image features $\boldsymbol{I}_{1: T}$, text features $\boldsymbol{C}_{1: T}$ and audio features $\boldsymbol{A}_{1: T}$ from speech, respectively. (b) The cross-modal alignment module then aligns these audio-image-text features at temporal and semantic levels. (c) The PMMTalk decoder driven by the image features $\boldsymbol{I}_{1: T}$, text features $\boldsymbol{C}_{1: T}$, audio features $\boldsymbol{A}_{1: T}$ and personal style $p$ predicts lip-syncing facial blendshape coefficients while controlling personal styles. These coefficients can be used to drive any digital characters according to the rendering engine rules.}
\label{method}
\end{figure*}

The perception of human speech fundamentally involves the integration of multiple modality signals~\cite{dupont2000audio,fan2022joint,mcgurk1976hearing}. Therefore, an exclusive reliance on single modality data could potentially lead to undesirable results. To bypass this barrier, as shown in Fig.~\ref{method}, we propose a novel architecture, namely PMMTalk, utilizing pseudo multi-modal features to generate accurate and synchronized lip movement. Let $\mathcal{X}_{1: T}=\left(\mathcal{x}_{1}, \ldots, \mathcal{x}_{T}\right)$ be a sequence of speech snippets, and each $\mathcal{x}_{t} \in \mathbb{R}^{D}$ has $\boldsymbol{D}$ samples to align to the corresponding (visual) frame $\boldsymbol{b}_{t}$. Let $\boldsymbol{B}_{1: T}=\left(\boldsymbol{b}_{1}, \ldots, \boldsymbol{b}_{T}\right)$, $\boldsymbol{b}_{t} \in \mathbb{R}^{32}$ be a $T$-length sequence of facial blendshape coefficients, which describes the ground-truth of 3D face movements. The proposed model takes speech $\mathcal{X}_{1: T}$, a random reference face image $R$ (prior structural information), and personal style $p \in \mathbb{R}^{64}$ as input, predicting facial blendshape coefficients $\hat{\boldsymbol{B}}_{1: T}=\left(\hat{\boldsymbol{b}}_{1}, \ldots, \hat{\boldsymbol{b}}_{T}\right)$. Formally,
\begin{equation}
\label{PMMTalk}
\hat{\boldsymbol{b}}_{t}=\operatorname{PMMTalk}_{\theta}\left(\mathcal{x}_{t}, R, p\right),
\end{equation}
where $\theta$ indicates the model parameters, $t$ is the current time-step in the sequence and $\hat{\mathbf{b}}_{\mathbf{t}} \in \hat{\mathbf{B}}_{\mathbf{T}}$. For the remainder of this section, we will provide a description of each PMMTalk component in detail.

\subsection{PMMTalk Encoder}
\subsubsection{Audio Embedding}
The design of audio embedding adopts the state-of-the-art self-supervised pre-trained speech model, wav2vec 2.0~\cite{baevski2020wav2vec}. By training on a sufficient amount of audio data, it can effectively extract the desired audio features, ensuring that the extracted features contain comprehensive information about the spoken words. Specifically, the audio embedding is bifurcated into two integral components: an audio encoder denoted as $\psi_{AE}(\cdot)$ and an audio decoder represented as $\psi_{AD}(\cdot)$. The former, comprising multiple layers of temporal convolutional networks, transforming the raw speech $\mathcal{X}$ input into latent speech representations $\mathbf{L}_{1: T}=\left(\boldsymbol{l}_{1}, \ldots, \boldsymbol{l}_{T}\right)$. On the other hand, the latter, which incorporates a stack of multi-head self-attention~\cite{vaswani2017attention} and feed-forward layers, transposing these latent speech feature vectors into contextualized speech representations $\boldsymbol{A}_{1: T}=\left(\boldsymbol{a}_{1}, \ldots, \boldsymbol{a}_{T}\right)$. The formula is defined as follows:
\begin{equation}
\label{Audio Embedding}
\mathbf{L}_{1: T}=\psi_{AE}(\mathcal{X}),
\boldsymbol{A}_{1: T}=\psi_{AD}(\mathbf{L}_{1: T}),
\end{equation}
where $T$ is the frame length. Here, we freeze the parameters of $\psi_{AE}(\cdot)$ while training. 

\subsubsection{Text Embedding}
The purpose of text embedding is to identify and extract informative latent features embedded within the input audio signal, thereby generating the corresponding textual information. A speech recognizer module, wav2vec 2.0~\cite{baevski2020wav2vec}, is leveraged to pursue this goal. In this approach, the input audio signal is initially transformed into waveform embeddings, and then mask and predict specific portions. This method efficiently captures and translates comprehensive audio features into text.

The text embedding is specifically composed of two components: an audio encoder $\psi_{AE}(\cdot)$ and a text decoder $\psi_{TD}(\cdot)$. The audio encoder $\psi_{AE}(\cdot)$ shares weights with the one in the audio embedding module. The text decoder $\psi_{TD}(\cdot)$, on the other hand, comprises 24 transformer blocks~\cite{vaswani2017attention} and a pre-trained fully connected layer. This structure maps the 1024-dimensional latent speech feature vectors on a frame-by-frame basis into textual vocabulary representations $\boldsymbol{C}_{1: T}=\left(\boldsymbol{c}_{1}, \ldots, \boldsymbol{c}_{T}\right)$ with 3503 dimensions. Formally:
\begin{equation}
\label{Text Encoder1}
\boldsymbol{C}_{1: T}=\psi_{TD}(\boldsymbol{L}_{1: T}),
\end{equation}
where $T$ is the frame length. Take note that the parameters of the text embedding module remain frozen during the training process.

\subsubsection{Image Embedding}
The image embedding module comprises two components: the talking head generator $\psi_{HG}(\cdot)$ and the image encoder $\psi_{IE}(\cdot)$. The talking head generator $\psi_{HG}(\cdot)$ utilizes the widely recognized architecture, Wav2Lip~\cite{prajwal2020lip}, to generate lip-syncing visual information. It specifically includes an identity encoder, a speech encoder, and a face decoder, all composed of stacks of 2D convolutional layers with residual skip connections. It requires an audio sequence $\mathcal{X}$ and a random reference face image $R$ providing prior structural information as inputs and outputs the corresponding lip movements video $\mathbf{V}_{1: T}=\left(\boldsymbol{v}_{1}, \ldots, \boldsymbol{v}_{T}\right)$. Sequentially, $\mathbf{V}_{1: T}$ is fed into image encoder $\psi_{IE}(\cdot)$ to generate image features $\boldsymbol{I}_{1: T}=\left(\boldsymbol{i}_{1}, \ldots, \boldsymbol{i}_{T}\right)$, which consists of a weight-sharing 2D CNN network and a transformer encoder~\cite{vaswani2017attention} with multiple layers, capturing short-term and long-term dependencies in the video. Formally, we define:
\begin{equation}
\label{Text Encoder}
\boldsymbol{V}_{1: T}=\psi_{HG}(\mathcal{X}, R),
\boldsymbol{I}_{1: T}=\psi_{IE}(\mathbf{V}_{1: T}),
\end{equation}
where $T$ is the frame length. Please note that the parameters of the talking head generator $\psi_{HG}(\cdot)$ stay frozen throughout the training process.

\subsection{Cross-modal Alignment}
Given that acoustic, visual, and textual features are different modalities, each with unique characteristics, it is crucial to discover their inherent mutual connections for effective collaboration. As these features share the same semantic information at the sentence level and reflect the same lip movement frame-by-frame in timing, an alignment at temporal and semantic levels is necessary.

\textbf{Temporal level.} Inspired by CLIP~\cite{radford2021learning}, we handle temporal alignment in a similar fashion. Specifically, we use $\left(\boldsymbol{X}_{1:T}, \boldsymbol{Y}_{1:T}\right)$ to denote the pair of different modality combinations including $\left(\boldsymbol{I}_{1:T}, \boldsymbol{C}_{1:T}\right)$, $\left(\boldsymbol{I}_{1:T}, \boldsymbol{A}_{1:T}\right)$, and $\left(\boldsymbol{C}_{1:T}, \boldsymbol{A}_{1:T}\right)$. Initially, these features are fed into a linear layer, with layer normalization (LN) subsequently applied to maintain a consistent scaling. Formally,
\begin{equation}
\label{Temporal Loss1}
\left(\hat{\boldsymbol{X}}_{1: T}, \hat{\boldsymbol{Y}}_{1:T}\right)={\text{LN}}\left({\text{Linear}}\left(\boldsymbol{X}_{1:T}, \boldsymbol{Y}_{1:T}\right)\right).
\end{equation}
The pair $\left(\hat{\boldsymbol{X}}_{1: T}, \hat{\boldsymbol{Y}}_{1:T}\right)$ is then pairwise scaled and evaluated for similarity. Formally,
\begin{equation}
\label{Temporal Loss-2}
{\boldsymbol{D}}_{1: T}={\text{Scale}}(\hat{\boldsymbol{X}}_{1: T}\times\hat{\boldsymbol{Y}}_{1: T}^\top).
\end{equation}
Following this, we use a Kullback-Leibler divergence loss function to determine the corresponding loss value.

\textbf{Semantic level.} Furthermore, despite their distinctive characteristics, the feature pair of different modality  $\left(\hat{\boldsymbol{X}}_{1:T}, \hat{\boldsymbol{Y}}_{1:T}\right)$ exhibit shared semantic information. Therefore, aligning them at the sentence-level assists in maintaining semantic consistency. Specifically, we initially apply an averaging operation to each pair. Formally,
\begin{equation}
\label{Temporal Loss}
\left(\tilde{\boldsymbol{X}}, \tilde{\boldsymbol{Y}}\right)={\text{Average}}\left(\hat{\boldsymbol{X}}_{1:T}, \hat{\boldsymbol{Y}}_{1:T}\right).
\end{equation}
Subsequently, we endeavor to maximize the cosine similarity of each corresponding pair, aligning these features within the same semantic space.

\subsection{PMMTalk Decoder}
To generate the 3D facial blendshape coefficients for lip-syncing, we utilize a module inspired by the Transformer decoder~\cite{vaswani2017attention}. PMMTalk decoder $\psi_{PD}(\cdot)$ consists of multiple layers of masked self-attention and feed-forward neural networks. The masked self-attention mechanism enables the decoder to focus on relevant parts of the input sequence to produce suitable outputs. 

Specifically, we first fed the image feature $\boldsymbol{I}_{1:T}$, the text feature $\boldsymbol{C}_{1:T}$, the audio feature $\boldsymbol{A}_{1:T}$ into a linear layer and then concatenate these with the personal style features $p$ into the decoder, which control the unique characteristics of facial expressions. Then they are encoded with a periodic positional encoding ~\cite{fan2022faceformer}, which captures the consistent timing of lip movements during speech, specifically when the mouth opens and closes. Following this, a biased multi-head self-attention layer is applied, integrating the positional encoding into the multi-head attention layers, taking inspiration from attention with linear biases~\cite{press2021train}. This layer assigns greater importance to closer information in the mask layer, focusing on the transitions between adjacent actions. Subsequently, the masked self-attention layer calculates a weighted sum of the input features based on their relevance to each other. Finally, a feed-forward neural network and a fully connected layer are utilized to produce facial blendshape coefficients.

\subsection{Loss function}
In order to train our neural network, we utilize a comprehensive loss function that encompasses four key components: position loss, motion loss, temporal loss, and semantic loss. The complete function is defined as follows:
\begin{equation}
\label{loss_all} 
L=\lambda_{1} L_{\text{pos}}+\lambda_{2} L_{\text{mot}}+\lambda_{3} L_{\text{tem}}+\lambda_{4} L_{\text{sem}},
\end{equation}
where $\lambda_{1}$ = 1, $\lambda_{2}$ = 10, $\lambda_{3}$ = $10^{-4}$, and $\lambda_{4}$ = $10^{-5}$ in all of our experiments. A comprehensive explanation of each of these components is provided in the following section.

\textbf{Position Loss.} The position loss measures the difference between the predicted facial blendshape coefficients and the corresponding ground-truth facial blendshape coefficients. Specifically, we use per-frame mean squared error
(MSE) as the position loss:
\begin{equation}
\label{loss_pos} 
L_{\text{pos}}=\frac{1}{T}\sum_{\mathbf{t}=1}^{\mathbf{T}}\left\|\left(\hat{b}_{t} -{b}_{t}\right)\right\|^{2}.
\end{equation}

\textbf{Motion Loss.} To address jittery output frames, we introduce a motion loss that ensures temporal stability. This loss considers the smoothness between predicted and ground truth frames:
\begin{equation}
\label{loss_mot} 
L_{mot}=\frac{1}{T}\sum_{\mathbf{t}=1}^{\mathbf{T}}\left\|\left(\hat{b}_{t}-\hat{b}_{t-1}\right)-\left(b_{t}-b_{t-1}\right)\right\|^{2}.
\end{equation}

\textbf{Temporal Loss.} The temporal loss is specifically designed to ensure the alignment of individual modal information at the frame level. Specifically, let ${\boldsymbol{W}}_{1:T}$ be a diagonal matrix with a diagonal filled with 1 as the label. Subsequently, we respectively fed the prediction and label into LogSoftmax and Softmax layers. Following this, we compute the temporal loss:

\begin{small}
\begin{equation}
\begin{split}  
\tilde{\boldsymbol{D}}_{1: T}={\text{LogSoftmax}}\left({\boldsymbol{D}}_{1: T}\right), \tilde{\boldsymbol{W}}_{1:T}={\text{Softmax}}\left({\boldsymbol{W}}_{1:T}\right),\\
L_{tem}=-\frac{1}{3}\sum(\tilde{\boldsymbol{D}}_{1: T}\log(\tilde{\boldsymbol{W}}_{1:T})+\tilde{\boldsymbol{W}}_{1:T}\log(\tilde{\boldsymbol{D}}_{1: T})).
\end{split}
\end{equation}
\end{small}

\textbf{Semantic Loss.} The semantic loss is designed to align the different modal information at a high-dimensional semantic level. To achieve this, we calculate the cosine similarity between each different modal pair $(\tilde{\boldsymbol{X}}, \tilde{\boldsymbol{Y}})$. Formally, we define:
\begin{equation}
\label{loss_sem} 
L_{sem}=\frac{1}{3}\sum(1-\frac{\tilde{\boldsymbol{X}}^{\top} \cdot \tilde{\boldsymbol{Y}}}{\|\tilde{\boldsymbol{X}}\|\|\tilde{\boldsymbol{Y}}\|}).
\end{equation}

\subsection{Experimental Setup}
\textbf{Training details.} 
Within the preliminary pre-processing phase, we first convert the audio sampling rate to 16 kHz, subsequently aligning the facial animation to correspond accurately with the audio. The VOCASET~\cite{voca} and the 3D-CAVFA datasets are processed at a rate of 30fps. Our primary objective is the enhancement of facial animation accuracy, we only select 32-dimensional facial blendshape coefficients. These coefficients, bearing a significant connection with pronunciation, serve as our training labels.

Throughout the training phase dedicated to model optimization, the model undergoes an end-to-end optimization process using the Adam optimizer ~\cite{kingma2014adam}. The learning rate and batch size are set to 1e-4 and 1, respectively. The entire network was trained on four NVIDIA GTX 409, and it took around 8 hours (80 epochs) to complete the training process.

\textbf{Datasets.} In this research, we utilize two 3D talking face datasets, VOCASET~\cite{voca} and 3D-CAVFA, for both training and testing. VOCASET~\cite{voca} comprises 480 high frame-rate (60fps) face motion sequences of 12 subjects, each spanning 3 to 4 seconds. Moreover, each 3D face mesh contains 5023 vertices. The 3D-CAVFA includes facial blendshape coefficients synchronized with facial movements for 20 subjects and spans a total duration of almost 15.3 hours. Covering commonly used words, phrases, and sentence structures in modern Chinese, the 3D-CAVFA dataset is highly applicable to daily situations.

\textbf{Rendering Engine.} In this paper, we utilize Unreal Engine (UE), a broadly utilized game development engine, for constructing our digital avatar. It allows us to separate animation from the character, enabling the pre-trained speech-driven 3D facial animation model to drive any digital character following the guidelines. This gives us a reliable and character-independent platform that necessitates only minimal prior knowledge of game creation.

\textbf{Baseline Implementations.} We measured the performance of our method, PMMTalk, against state-of-the-art methods including VOCA~\cite{voca}, MeshTalk~\cite{meshtalk}, FaceFormer~\cite{fan2022faceformer}, CodeTalker~\cite{xing2023codetalker}, and EmoTalk~\cite{emotalk}. For the 3D-CAVFA dataset, which is in Chinese, performance variances could significantly impact these previously trained English models. Uniquely, the 3D-CAVFA dataset uses facial blendshape coefficients as labels, differing from the official datasets which employ the vertex offsets of mesh as labels. To ensure fairness, we adapted the audio feature extractor for all models with the XLSR-53~\cite{conneau2020unsupervised}, a wav2vec 2.0~\cite{baevski2020wav2vec} model pre-trained on data in 53 languages. Furthermore, we modified the final linear classification layer of these models to match the 3D-CAVFA dataset.

For the VOCASET dataset, we employed the provided pre-trained weights for testing on VOCA~\cite{voca}, FaceFormer~\cite{fan2022faceformer}, and CodeTalker~\cite{xing2023codetalker}. For MeshTalk~\cite{meshtalk}, we used its official implementation for both training and testing. For EmoTalk~\cite{emotalk} and PMMTalk, we made adjustments to the last linear classification layer to fit the VOCASET set. All methods, except for MeshTalk, necessitate conditioning on a training speaking style during testing. When dealing with unseen subjects, we generate facial animations by conditioning on all available training styles.

\section{Experiments}
\begin{table*}[!htbp]
\begin{minipage}{0.55\textwidth}
\centering
\caption{Quantitative evaluation results on different protocol of 3D-CAVFA. Lower means better for both metrics.} 
\setlength{\tabcolsep}{6pt}
\renewcommand\arraystretch{1.5} 
\begin{tabular}{c|c|c|c|c|c}
                \hline
               \multirow{3}{*}{Method} &\multirow{3}{*}{Ref.} &\multicolumn{2}{c|}{cross-subject} &\multicolumn{2}{c}{cross-gender} \\
               \cline{3-6} 
                                            &&LVE$\downarrow$ 
                                            &ALE$\downarrow$ 
                                            &LVE$\downarrow$ 
                                            &ALE$\downarrow$\\
                &&($\times10^{-2}$)&($\times10^{-2}$)&($\times10^{-2}$)&($\times10^{-2}$)\\
                \hline 
                VOCA~\cite{voca}                            &CVPR'19 &9.866 &0.950 &7.478 &0.713\\
                \hline
                FaceFormer~\cite{fan2022faceformer}         &CVPR'22 &7.350 &0.642 &7.762 &0.673\\
                \hline
                CodeTalker~\cite{xing2023codetalker}        &CVPR'23 &13.206 &1.310 &11.385 &0.000\\
                \hline
                EmoTalk~\cite{emotalk}                      &ICCV'23 &7.591 &0.684 &7.096 &0.636\\
                \hline
                \textbf{PMMTalk(Ours)} &- &\textbf{7.169} &\textbf{0.640} &\textbf{6.568} &\textbf{0.590}\\
               \hline

\end{tabular}
\label{Quantitative-3DCAVFA}
\end{minipage}
\hfill
\begin{minipage}{0.45\textwidth}
\centering
\caption{Quantitative evaluation results on VOCA-Test. Lower means better for LVE metric.} 
\setlength{\tabcolsep}{12pt}
\renewcommand\arraystretch{1.5} 
\begin{tabular}{c|c|c}
                \hline
                \multirow{2}{*}{Method} & \multirow{2}{*}{Ref.} & LVE$\downarrow$ \\
                &&($\times10^{-5}$mm)\\
                \hline 
                VOCA~\cite{voca}                           &CVPR'19 &4.925\\
                \hline
                MeshTalk~\cite{meshtalk}                   &ICCV'21 &4.544\\
                \hline
                FaceFormer~\cite{fan2022faceformer}        &CVPR'22 &4.109\\
                \hline
                CodeTalker~\cite{xing2023codetalker}       &CVPR'23 &3.945\\
                \hline
                EmoTalk~\cite{xing2023codetalker}          &ICCV'23 &3.536\\
                \hline
                \textbf{PMMTalk(Ours)}                   &- &\textbf{2.990}\\
                \hline
\end{tabular}
\label{Quantitative-voca}
\end{minipage}
\end{table*}

\subsection{Quantitative Evaluation}
\textbf{Lip-sync analysis.} To measure lip synchronization, we calculated the lip vertex error (LVE) as used in CodeTalker~\cite{xing2023codetalker}, which is computed as the average $l_2$ error of the lips in the test set. For a single frame, the LVE is defined as the maximum $l_2$ error among all lip vertices for the VOCASET dataset, or among the blendshape coefficients for the 3D-CAVFA dataset, respectively. Nevertheless, using the LVE metric may potentially be influenced by outliers present in the dataset. To attenuate the impact of outliers and provide a more holistic evaluation, we additionally calculated the lip average error. Here, we applied the Average Lip Error (ALE) to compute the average $l_2$ error of blendshape coefficients per frame and then averaged this across the test set, which reflects the overall perceived quality of mouth movement. 

As shown in Table~\ref{Quantitative-3DCAVFA}, the performance of different models is evaluated on both protocols. Firstly, it is evident that the overall values obtained from cross-gender protocol are lower than those from cross-subject protocol. This discrepancy may be attributed to the fact that cross-gender protocol benefits from a larger training set compared to cross-subject protocol. Moreover, compared to previous methods, our approach exhibits lower LVE and ALE values, indicating superior performance. Our model is capable of generating more accurate lip movements than the other models.

\textbf{Robustness analysis.} Moreover, as indicated in Table~\ref{Quantitative-voca}, we embarked on a comprehensive comparison with previous methods on the VOCA dataset. For the sake of fairness, we calculated the same metric (LVE) used in \cite{meshtalk, fan2022faceformer, xing2023codetalker}. Through this evaluation, we trained our model on VOCASET and then tested it on VOCA-Test. Firstly, the models exhibit varied performance on VOCA-Test and 3D-CAVFA, a divergence that may arise from the distinct labels of datasets. The former employs vertex offsets of the mesh (5023*3), while the latter relies on the weight values of a 32-dimensional facial blendshape. Moreover, the numerical results confirm that our approach achieves a lower lip error, which demonstrates the superior performance of our proposed method over multiple datasets, substantiating its effectiveness and robustness in comparison to existing methods.

\begin{figure}[htbp]
\centering
\includegraphics[width=9cm]{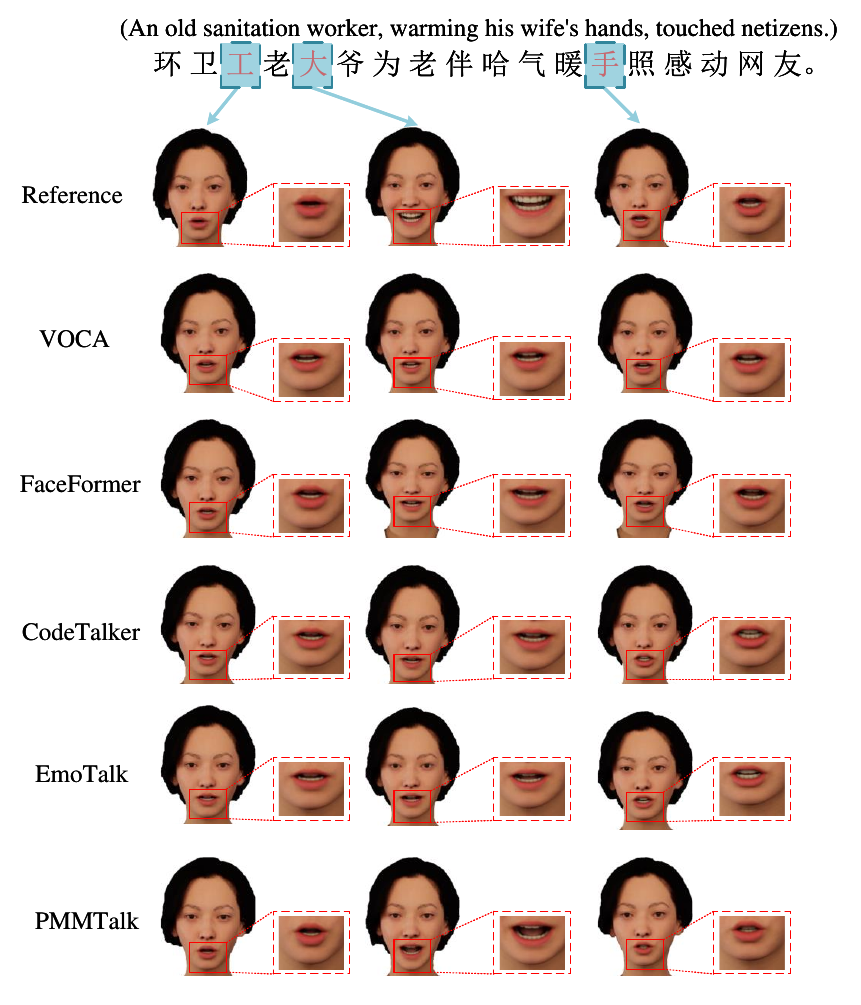}
\caption{\textbf{Qualitative comparison of facial movement by different methods on 3D-CAVFA.} We compared different words and found that our method yields performance close to that of the ground truth, producing more precise mouth movements compared to other models.}
\label{Qualitative Evaluation}
\end{figure}

\subsection{Qualitative Evaluation}
While quantitative metrics are indispensable for evaluating 3D talking faces, the visualization of prediction results is equally critical for a holistic understanding of the model's performance. Consequently, we implemented qualitative evaluation against all models, ensuring fair assessment by assigning the same speaking style as the conditional input. We suggest viewing our supplementary video for the prediction performance of multiple methods. Specifically, the content of the video is as follows: (1) audio sequences from 3D-CAVFA test sets in the cross-subject protocol; (2) in-the-wild audio clips extracted from China meteorological observatory; (3) different language audio clips extracted from supplementary videos of previous methods. The video reveals that PMMTalk generates lip semantic movements that are not only more coherent and realistic but also enhance lip movement comprehensibility. Furthermore, PMMTalk exhibits impressive performance across multiple languages. Additionally, we also demonstrated different speaking styles for the same sentence.

Additionally, we assessed the lip synchronization performance and depicted three typical frames of synthesized facial animations speaking specific words, as shown in Fig.~\ref{Qualitative Evaluation}. A comparison reveals that our PMMTalk generates lip movements that are not only more accurate with respect to the corresponding speech signals but are also more in alignment with the ground truth. For example, when articulating the word $/gong/$, PMMTalk achieves optimal lip-syncing with an appropriate mouth pout, echoing the required rounding and slight elevation of the lips. 
In a similar vein, when pronouncing the word $/shou/$, the lips naturally part, shaping into a circular form while gently curling upwards. Furthermore, our method presents a more appropriate movement with a full mouth opening, downward jaw motion, and the corner of the mouth extending sideways while pronouncing the word $/da/$. These visualizations clearly underscore that our method outperforms previous techniques in terms of generating more accurate lip shapes.

\begin{table}[!htbp]
\centering
\caption{\textbf{User study results.} We established two sub-tasks, specifically lip movement realism (\#realism) and perceptual lip synchronization (\#lip-sync). The Mean Opinion Score (MOS) for these two categories is reported.}
\setlength{\tabcolsep}{10pt}
\renewcommand\arraystretch{1.5} 
\begin{tabular}{c|c|c|c}
                \hline
                Method & {Ref.} & \#realism$\uparrow$ & \#lip-sync$\uparrow$  \\
                \hline 
                VOCA~\cite{voca}                           &CVPR'19 &1.54 &1.49\\
                \hline
                CodeTalker~\cite{xing2023codetalker}       &CVPR'22 &2.59 &2.91\\
                \hline
                FaceFormer~\cite{fan2022faceformer}        &CVPR'23 &3.61 &3.70 \\
               \hline
                Emotalk~\cite{emotalk}                     &ICCV'23 &3.36 &3.57\\
               \hline
                \textbf{PMMTalk(Ours)}                  &- &\textbf{4.27} &\textbf{4.49}\\
                \hline
\end{tabular}
\label{userstudy}
\end{table}

\subsection{User study}
A user study is a reliable method for evaluating generated 3D talking faces. In this section, we conducted a user study to further compare our model against previous ones and the ground truth. The human perceptual system has been innately wired to capture and understand subtle facial motions, including lip synchronization. Therefore, we incorporated two metrics in the user study: lip movement realism and perceptual lip synchronization. The first metric evaluates the movement and smoothness of key facial features, including the mouth, jaw, and cheek. It examines the naturalness and fluency of the digital person's speech, assessing whether there is a mechanical feeling and unnecessary speech jittering. The latter metric primarily assesses the accuracy of the digital person's pronunciation in terms of correct mouth shapes and the seamless synchronization of lip movements with the audio. 

We devised a questionnaire consisting of 20 questions to evaluate the outcomes and invited 20 students to participate. Each question prompted participants to rate the Mean Opinion Score (MOS) for videos on a scale of 1-5, with a higher score indicating better results. For each question, the left video represented the ground truth, while the right video sequences depicted the outcomes of different models. The audio clips were sourced from the 3D-CAVFA test dataset of cross-subject protocol, encompassing words, phrases, and sentences. Participants were encouraged to evaluate each result based on their own subjective criteria. We specifically instructed them to concentrate on the area around the mouth, including lip, jaw, and cheek movements, while disregarding any unnatural effects related to the eyes.

\begin{table}[!htbp]
\centering
\caption{\textbf{Ablation study for our components.} We show the LVE and ALE in different cases.  Lower means better for both metrics.} 
\setlength{\tabcolsep}{18pt}
\renewcommand\arraystretch{1.5} 
\begin{tabular}{c|c|c}
\hline
\multirow{2}{*}{Method}     &LVE$\downarrow$ &ALE$\downarrow$\\
&($\times10^{-2}$)&($\times10^{-2}$)\\
\hline
Ours       &\textbf{7.169} &\textbf{0.640} \\
\hline
w/o Text Embedding     &7.319 &0.682\\
w/o Audio Embedding     &8.313 &0.786 \\
w/o Image Embedding     &7.459 &0.697 \\
\hline
w/o $L_{\text{tem}}$ Loss  &7.680 &0.704 \\
w/o $L_{\text{sem}}$ Loss  &7.599 &0.698 \\
w/o $L_{\text{mot}}$ Loss  &7.363 &0.681 \\
\hline
\end{tabular}
\label{ablation}
\end{table}

The scores pertaining to lip synchronization and realism are compiled in Table~\ref{userstudy}, indicating a preference for PMMTalk among the participants over its competitors. We infer that this preference is due to the facial animations synthesized by PMMTalk exhibiting more expressive facial movements, accurate lip shapes, and well-coordinated mouth movements. For instance, we outperformed VOCA by a margin of 2.73 points in the realism of lip movement and surpassed Faceformer by 0.79 points in perceptual lip synchronization. In totality, the user study substantiates that the facial animations generated by PMMTalk possess superior perceptual quality.

\subsection{Ablation experiment}
In this section, we conducted an ablation study to meticulously examine the contributions of various components within our model. Specifically, we delved into the repercussions of removing distinct feature embedding modules and employing different loss functions, aiming to discern their individual impacts on the overall model performance.

\textbf{Impact of different feature embedding modules.} In this section, whenever a particular feature embedding module is omitted, we continue to train the remaining modules. As depicted in Table~\ref{ablation}, the absence of the image embedding, text embedding, or audio embedding module often leads to a notable increase in both LVE and ALE, albeit at varying degrees. Notably, the exclusion of the audio embedding module causes the most significant impact, as evidenced by a substantial rise in the error rate of the lip shape for the generated facial animation. This is primarily because of the critical role the audio embedding module plays in human pronunciation. The outcomes from the other modules indicate that the ambiguity caused by the sole use of audio embedding modes can be reduced by addressing these with other modules.

\textbf{Impact of different loss function.} We carried out a thorough investigation into the impact of different losses by separately omitting them from our comprehensive model. The temporal loss guarantees alignment of individual modal information over time, whereas the semantic loss assures alignment of different modal information at a high-dimensional semantic level. Neglecting these losses resulted in a significant deterioration in the overall visual quality of the generated faces, as well as the clarity of lip movements. Particularly, eliminating the temporal loss resulted in a 0.511 increase in LVE and a 0.064 increase in ALE, while the omission of the semantic loss led to a surge of 0.430 in LVE and 0.058 in ALE. Additionally, we studied the impact of removing the motion loss, which encourages the predicted facial animations to closely match the movement velocity of the ground truth. Though omitting the motion loss resulted in a minor performance dip, it induced noticeable jitter in the output of facial animation. Accordingly, these outcomes indicate that motion loss can infuse prior facial motion based on ground truth labels, enabling the creation of more coherent facial animations.



\section{Conclusion}
\label{sec:conclusion}
In this study, we present PMMTalk, an innovative framework developed to enhance the accuracy of facial animations through the utilization of pseudo multi-modal features. Comprising three main components—the PMMTalk encoder, cross-modal alignment module, and PMMTalk decoder—the PMMTalk encoder employs the ready-made talking head generation method and speech recognition technology to extract visual and textual information from speech. Subsequently, the cross-modal alignment module aligns these audio-image-text features at both the temporal and semantic levels. The PMMTalk decoder utilizes these features to predict lip-synced facial blendshape coefficients. To tackle the limitation of insufficient 3D talking face data, we introduce a large-scale 3D talking face dataset, namely the 3D-CAVFA dataset, which is equipped with synchronized facial blendshape coefficients, a varied corpus, and a range of subjects. Experimental outcomes validate that our method outperforms prevailing state-of-the-art methods and garners more favorable user feedback. Our work stands to potentially contribute to virtual reality applications by enabling more authentic and emotionally engaging experiences with more lifelike talking faces.

\textbf{Limitations and Future Work.} Nevertheless, it's imperative to note that our method exclusively generates 32-dimensional facial blendshape coefficients, lacking the incorporation of broader elements such as facial expressions and head movements. Furthermore, our method depends on multiple large-scale pre-training models, which extends the inference time of models, thereby posing a challenge to real-time applications.
\printbibliography
\end{document}